\documentclass[conference]{IEEEtran}
\IEEEoverridecommandlockouts
\usepackage{cite}
\usepackage{amsmath,amssymb,amsfonts}

\usepackage{graphicx}
\usepackage{textcomp}
\usepackage{xcolor}
\usepackage{balance}

\usepackage{bm}             
\usepackage{algorithm}      
\usepackage{algpseudocode}  

\usepackage{subcaption}
\def\BibTeX{{\rm B\kern-.05em{\sc i\kern-.025em b}\kern-.08em
    T\kern-.1667em\lower.7ex\hbox{E}\kern-.125emX}}

\begin{document}

\title{\LARGE \bf 
Probabilistic Task Parameterization of Tool-Tissue Interaction via Sparse Landmarks Tracking in Robotic Surgery

\thanks{* Denotes equal contribution. $\dagger$ Denotes the corresponding author.}
\thanks{$^1$ Department of Electrical Engineering and Computer Science, University of Tennessee Knoxville, Knoxville, TN 37996, USA {\tt\small ywang306@vols.utk.edu, fliu33@utk.edu}.}%
\thanks{$^{2}$ Department of Mechanical Engineering, Stanford University, Stanford, CA 94305, USA {\tt\small yunxin6@stanford.edu}.}%
}

\author{Yiting Wang$^{1,*}$, Yunxin Fan$^{2,*}$, Fei Liu$^{1,\dagger}$%
\vspace{-2mm}
}

\maketitle

\begin{abstract}
Accurate modeling of tool-tissue interactions in robotic surgery requires precise tracking of deformable tissues and integration of surgical domain knowledge. Traditional methods rely on labor-intensive annotations or rigid assumptions, limiting flexibility. We propose a framework combining sparse keypoint tracking and probabilistic modeling 
that propagates expert-annotated landmarks across endoscopic frames, even with large tissue deformations. Clustered tissue keypoints enable dynamic local transformation construction via PCA, and tool poses, tracked similarly, are expressed relative to these frames. Embedding these into a Task-Parameterized Gaussian Mixture Model (TP-GMM) integrates data-driven observations with labeled clinical expertise, effectively predicting relative tool-tissue poses and enhancing visual understanding of robotic surgical motions directly from video data.

\end{abstract}


\section{Introduction}
Robotic-assisted minimally invasive surgery (RMIS) offers improved precision and reduced human error, yet tool-tissue interaction remains a core challenge due to tissue deformability, anatomical variability, and strict safety constraints. Traditional methods often depend on dense manual annotations or rigid kinematic assumptions, limiting scalability and generalization across surgical contexts. Recent learning-based approaches have improved tissue tracking, but typically overlook the integration of domain-specific surgical priors—crucial for interpretability and robustness in clinical settings.

To address this, we propose a hybrid framework that combines sparse keypoint tracking with probabilistic task parameterization. Using LocoTrack \cite{Seokju_2024}, we manually annotate and track sparse anatomical landmarks (e.g., tissue regions, tool tips) across video frames. These keypoints are clustered to generate dynamic tissue frames via PCA, capturing local deformation. Tool poses are expressed relative to these frames and modeled using a Task-Parameterized Gaussian Mixture Model (TP-GMM), which integrates both time and tissue context. Our formulation enables uncertainty-aware, context-sensitive trajectory prediction, incorporating surgical priors such as motion constraints and safety zones.
\begin{figure}[!htb]
    \centering
    \includegraphics[width=\linewidth]{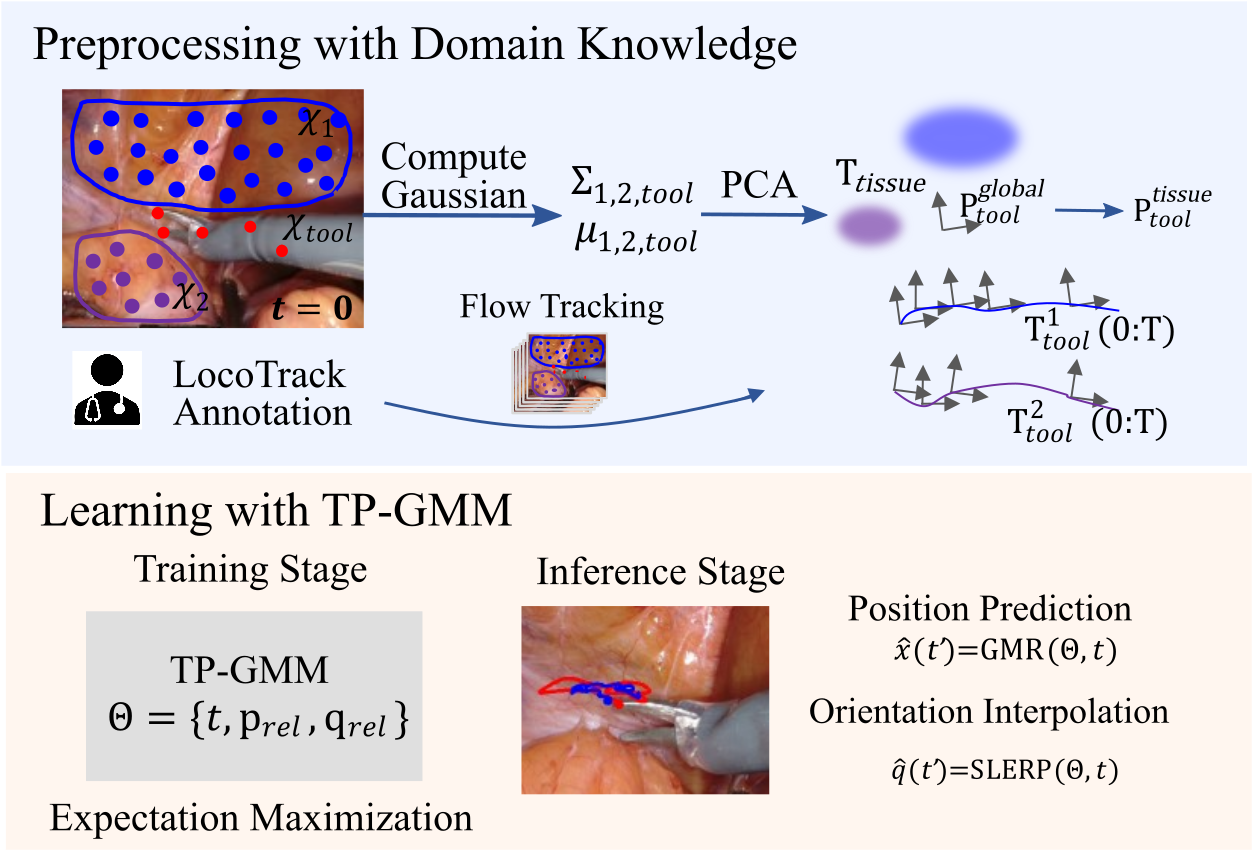}
    \caption{framework diagram}
    \label{fig:intro}
    \vspace{-5mm}
\end{figure}

\section{Method}

This section outlines our approach for modeling tool-tissue interactions using a time-conditioned TP-GMM, applied to the EndoNeRF dataset~\cite{wang2022neural}. We learn a model of tool motion in SE(2), capturing 2D position and orientation directly from endoscopic video. The method relies on sparse anatomical keypoints and dynamically derived local tissue frames. The overall pipeline is summarized in Algorithm~\ref{alg:tp_gmm_framework}, enabling smooth and context-aware motion prediction conditioned on surgical tool dynamics and tissue geometry.


\subsection{Framework Overview}

We model tool–tissue interactions using 2D keypoints tracked from endoscopic video via LocoTrack. Guided by surgical domain knowledge, we cluster tissue landmarks into semantically meaningful regions (e.g., tissue layers or boundaries), with the number and structure of clusters reflecting anatomical priors. Local tissue frames are then derived from these regions using PCA, and tool poses are expressed relative to the resulting dynamic reference frames. A task-parameterized Gaussian Mixture Model (TP-GMM) is trained on these frame-relative trajectories, enabling probabilistic modeling that captures spatiotemporal variability, uncertainty, and multi-modality in tool motion.

\subsection{Tissue-Relative Frame Representation}



A key challenge in surgical motion modeling lies in the nonrigid, time-varying nature of tissue. To address this, we construct local reference frames $\mathbf{T}_{\text{ref}}(t)$ from sparse tool and tissue landmarks, as detailed in Algorithm~\ref{alg:tp_gmm_framework}. Tool poses ${c(t), q(t)}$ are transformed into this frame, yielding a deformation-invariant representation. The resulting time- and tissue-conditioned trajectories are used to train a TP-GMM for smooth and context-aware motion prediction.



\begin{algorithm}[!htb]
\small
\caption{Tool-Tissue Interaction Modeling via Time-Conditioned TP-GMM}
\label{alg:tp_gmm_framework}
\begin{algorithmic}[1]
\State \textbf{Input:} Tracked tool landmarks $\mathcal{X}_{\text{tool}}(t)$ and tissue landmarks $\mathcal{X}_{\text{tissue}}(t)$ for $t = 1, \ldots, T$
\State \textbf{Output:} Predicted tool poses $\hat{x}_{\text{tool}}(t)$ for $t \in [0,1]$
\vspace{1mm}
\For{$t = 1$ to $T$}

    \State $\{\mathcal{C}_i(t)\} \gets \mathtt{DomainCluster}(\mathcal{X}_{\text{tissue}}(t))$, $i=1, \ldots, k$
    
    \State $\left(\mu_i(t), \Sigma_i(t)\right) \gets \mathtt{computeGaussian}(\mathcal{C}_i(t))$
    
    \State $c_{\text{ref}}(t) \gets 1/k \sum \mu_i(t)$
    \State $\mathbf{R}_{\text{ref}}(t) \gets \mathtt{PCA}(\{\mu_i(t)\})$
    \State $\mathbf{T}_{\text{ref}}(t) \gets \begin{bmatrix} \mathbf{R}_{\text{ref}}(t) & c_{\text{ref}}(t) \\ \mathbf{0}^\top & 1 \end{bmatrix}$ \Comment{{\footnotesize Reference frame}}
    
    \State $\left\{\mathbf{p}_{\text{rel}}(t),\ \mathbf{q}_{\text{rel}}(t)\right\} \gets \left\{\mathbf{T}_{\text{ref}}^{-1}(t) \cdot \begin{bmatrix} c(t) \\ 1 \end{bmatrix},\ q_{\text{ref}}^{-1}(t) \cdot q(t)\right\}$

    \State $\mathcal{P}(t) \gets \left[t, \mathbf{p}_{\text{rel}}(t),\mathbf{q}_{\text{rel}}(t)\right]$
\EndFor
\vspace{1mm}
\State $\Theta \gets \mathtt{TP\_GMM\_Train}(\mathcal{P}^\top)$ \Comment{{\footnotesize Train with EM}}
\vspace{1mm}
\For{$t \in [0,1]$}
    \State $\hat{x}(t) \gets \mathtt{GMR}(\Theta, t)$ \Comment{{\footnotesize Predict position}}
    \State $\hat{q}(t) \gets \mathtt{SLERP}(\Theta, t)$ \Comment{{\footnotesize Interpolate orientation}}
\EndFor
\vspace{1mm}
\State \Return $\{\hat{x}_{\text{tool}}(t)\} = \{[\hat{x}(t), \hat{q}(t)]\}$
\end{algorithmic}
\end{algorithm}
\vspace{-1mm}

\subsection{Learning with TP-GMM and Prediction}

To model temporal tool motion, we train a time-conditioned TP-GMM on frame-relative trajectories $\mathcal{P}(t)$ via Expectation Maximization. Each component is conditioned on time and tissue context, enabling the model to capture uncertainty and multimodal patterns in surgical demonstrations. In the inference stage, the tool positions $\hat{x}(t)$ are predicted using Gaussian mixture regression (GMR), and the orientations $\hat{q}(t)$ are computed using spherical linear interpolation (SLERP), ensuring smooth and consistent pose predictions. The resulting trajectories support the execution of tasks that adapt to robotic cutting or suturing.

\section{Results}

We evaluate the proposed time-conditioned TP-GMM on the EndoNeRF dataset for a 2D cutting task. The model is trained on 128 frames and tested on the remaining 28 frames. Performance is quantified by Euclidean distance between predicted and ground-truth positions. Orientation is assessed using angular errors. The model achieves smooth and visually consistent pose predictions over time.
\balance
\begin{figure}[!htb]
    \centering
    \includegraphics[width=\linewidth]{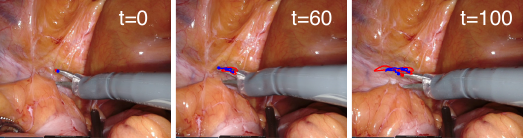}
    \caption{Comparison of ground truth (red) and predicted (blue) tool-tip trajectories across three frames.}
    \label{fig:video_frames}
    \vspace{-6mm}
\end{figure}


\subsection{Quantitative Evaluation}

We evaluated TP-GMM performance using Euclidean distance for position and angular difference for orientation. These metrics reflect how well the model fits the training data and generalizes to unseen deformations.

\subsubsection{Position Error}

The accuracy of the tool position is evaluated using the Euclidean distance between the predicted and ground-truth 2D pixelwise positions (in pixel). The mean errors for the training and test set are 12.97 px and 67.69 px, respectively. Notice that the larger testing error indicates overfitting and limited generalization to unseen tissue motions.

\subsubsection{Angle Error}

Orientation is evaluated using the angular error (in degrees), calculated as the absolute difference between the predicted and ground-truth tool angles. This metric is better suited than quaternion-based metrics in our 2D planar setup. The mean angular errors for the training and test set are 1.93° and 5.48°, respectively. This trend mirrors the position error, showing strong training performance but reduced accuracy on the test set.

\begin{figure}[!htb]
    \centering
    \includegraphics[width=\linewidth]{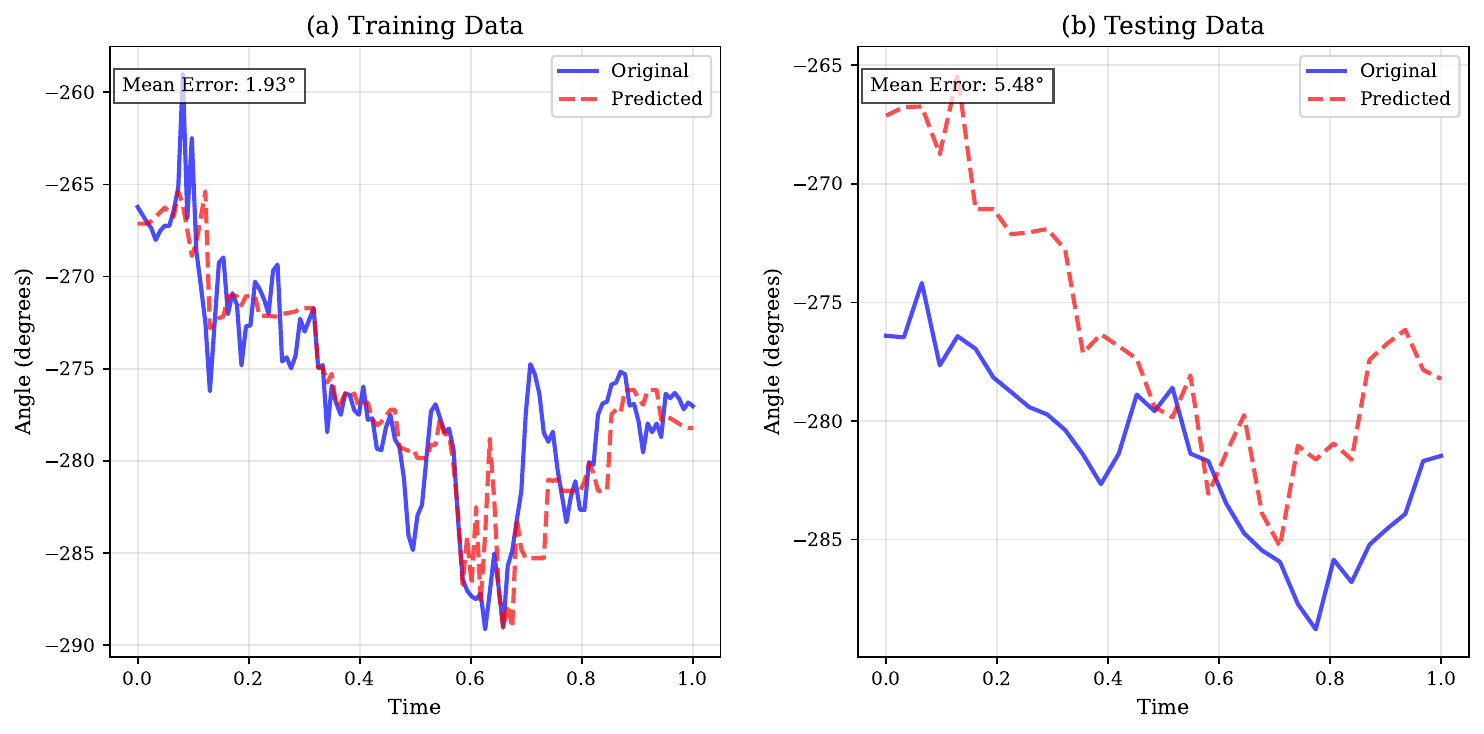}
    \caption{Angular error for training (a) and testing (b)}
    \label{fig:angle_analysis}
    \vspace{-6mm}
\end{figure}

Figure~\ref{fig:angle_analysis} presents angle error curves in normalized time (0–1) for both training and testing sets, highlighting how temporal dynamics affects prediction accuracy.

    

    

\section{Discussion}

The results highlight both strengths and limitations of the TP-GMM framework. Low training error (12.97 px) confirms its ability to model tool–tissue interactions, with GMR and SLERP ensuring smooth position and orientation predictions. However, high test error (67.69 px) and a 5.2× test/train ratio indicate overfitting, likely due to limited data diversity and over-parameterization from adaptive mixture components ($N = 10 \sim 30$).

Future work will explore data augmentation, regularization, and expanded training sets to improve generalization. Incorporating quantitative orientation metrics (e.g., angular distance) may further enhance evaluation. Despite current limitations, TP-GMM shows strong potential for surgical assistance and downstream use in Learning from Demonstration (LfD) and Imitation Learning (IL) frameworks.



\bibliographystyle{IEEEtran}
\bibliography{main}

\end{document}